\documentclass[12pt]{article}

\usepackage[style=apa, backend=biber, sortcites=false]{biblatex}
\usepackage{graphicx}
\usepackage{amssymb}
\usepackage{multirow}
\usepackage{amsmath}
\usepackage[all]{xy}
\usepackage{tikz}

\DeclareMathOperator*{\argmax}{argmax}
\usepackage{amsthm}
\theoremstyle{definition} 
\newtheorem{assumption}{Assumption}
\newtheorem{definition}{Definition}
\newtheorem{theorem}{Theorem}

\begin{document}

\noindent Arvid Sj\"olander\\
Karolinska Institutet\\
arvid.sjolander@ki.se

\LARGE
\begin{center}
A causal modeling perspective on decision theory
\end{center}
\normalsize

\section*{Abstract}

Decision theory provides a formal framework for how agents should make choices under uncertainty, drawing on ideas from philosophy, probability, and causality. Despite significant progress, the field still lacks a unified modeling language, and key concepts -- such as the distinction between subjective and objective elements, or what it means for a decision theory to \emph{perform well} -- are often left implicit. This can make it difficult to evaluate and compare competing theories, particularly in controversial cases. In this paper, we address these issues by introducing a formal framework for decision theory based on nonparametric structural equation models (NPSEMs), a well-established tool in causal inference. NPSEMs provide a unified foundation for representing agents, counterfactuals, and causal relationships, allowing for unambiguous definitions of EDT and CDT. Building on this foundation, we propose a novel decision theory --  personal decision theory -- which instructs agents to maximize a subjective model of their own counterfactual utility. We introduce a formal performance metric based on hypothetical interventions that enforce a given decision theory across a population -- such as might be achieved through education or policy -- and show that, under certain assumptions, personal decision theory is optimal with respect to this metric. Throughout, we use the smoking lesion problem as a running example and conclude with a formal analysis of Newcomb’s problem. Our aim is to provide decision theory with a clearer modeling language and firmer evaluative ground, thereby enabling more rigorous comparisons and facilitating conceptual progress in the field.

\section{Introduction}

Decision theory is a cross-disciplinary field that draws upon concepts from philosophy, probability theory, statistics, logic, and epistemology to model how agents should make decisions under uncertainty. At its core, decision theory formalizes the situation where an agent must choose between possible `acts', each of which may lead to outcomes with different `utility' \parencite{peterson2017introduction,sep-decision-theory}. 

Among the competing schools of thought within decision theory, two stand out as particularly influential: evidential decision theory (EDT) and causal decision theory (CDT). Both prescribe that agents choose the act with the highest expected utility, but differ in how they define this expectation. EDT evaluates an act based on how strongly it correlates, in a statistical sense, with desirable outcomes. CDT, on the other hand, asks what causal effect the act will have on the outcome. 

While these theories sometimes agree, they often do not. In particular, they tend to prescribe different acts in certain `exotic' problems, such as Newcomb’s problem, Death in Damascus, and the XOR blackmail problem, which typically involve a highly accurate predictor of the agents' acts. These problems have sparked decades of lively debate, with no consensus on which theory is most appropriate, or truly captures rational behavior.

One likely reason for this lack of resolution is that the literature on decision theory is quite fragmented: authors often use different notations, definitions, and formalisms, which complicates meaningful comparison and hinders progress. A compounding issue is that the models that authors use are often under-specified -- key concepts are left vague, or not defined at all. For example, authors 
 frequently claim that one decision theory outperforms another without defining what it means for a theory to \emph{perform well}.\footnote{For instance, referring to Newcomb's problem, 
 \textcite[][p. 151]{joyce1999foundations} stated: `I'm \emph{convinced} that causal decision theory has it right here' (italics added), without giving any formal motivation. Similarly, referring to Newcomb's problem and Death in Damascus, \textcite{levinstein2020cheating} asserted, also with no formal motivation, that `EDT gets these cases right, while CDT gets them wrong.'} In the absence of an explicit performance metric, such claims risk being little more than subjective opinions. Likewise, authors often refer to probabilities and expected utilities as being `subjective', yet rarely provide a formal distinction between subjective and objective elements.\footnote{For instance, \textcite{levinstein2020cheating} wrote: `We take for granted that the agent comes equipped with a ... credence function $P$ that measures her subjective degrees of confidence...', without commenting on how this subjective credence function relates to the objective reality. Similarly, \textcite{hitchcock2016conditioning} wrote: `The agent’s credence or subjective probability is represented by $P$, and explicitly added: `...I will not take pains to distinguish subjective and objective probability'.} Without such a distinction, it becomes unclear how different theories would fare under an objective performance metric -- even if such a metric were to be specified.

This ambiguity is particularly acute when it comes to causality. While counterfactuals have long served as a tool for expressing causal concepts in decision theory \parencite{lewis1973counterfactuals,gibbard1978counterfactuals}, they are notoriously difficult to interpret. As \textcite[][p. 100]{pearl2009causality} emphasizes: `When counterfactual variables are not viewed as by-products of a deeper, process-based model, it is hard to ascertain whether all relevant counterfactual independence judgments have been articulated, whether the judgments articulated are redundant, or whether those judgments are self-consistent'. This lack of clarity has had not only philosophical consequences but practical ones as well, as evidenced by known pitfalls in fields like medicine \parencite[][Section 11.3.5]{pearl2009causality}.

In contrast, causal diagrams offer a more intuitive and visually accessible modeling framework for causality, and have gained traction in decision theory in recent years \parencite[e.g.,][]{hitchcock2016conditioning,levinstein2020cheating}. However, their formal status in decision theory is often unclear; decision theorists tend to treat causal diagrams as as heuristic devices rather than formal mathematical objects, which raises concerns about the rigor and scope of the conclusions drawn from them.

To address these issues, we propose to use nonparametric structural equation models (NPSEMs) as a unifying foundation for decision theory. These models originate from linear structural equation models, which have been used in econometrics and social science for almost a century \parencite[see][and the references therein]{bollen1989structural}. NPSEMs were developed in a series of papers by Judea Pearl \parencite[see][and the references therein]{pearl2009causality} and provide a formal bridge between counterfactuals and causal diagrams, without making parametric assumptions about the problem at hand. They are widely accepted in modern statistics and causal inference; yet, their potential in decision theory appears to be unexplored.

We will show how to formally represent agents within the NPSEM framework, and we will suggest how to use this framework to provide unambiguous definitions of EDT and CDT. We will further propose a novel decision theory that we label `personal decision theory' (PDT). We will propose an objective performance metric that captures the notion of enforcing a decision theory uniformly on a population of agents and we will demonstrate that, with respect to this performance metric, PDT is optimal under certain assumptions. We will use the `smoking lesion problem' as a running example, which is a classic problem in decision theory. This particular problem appears to be relatively uncontroversial, which makes it suitable for pedagogical purposes. We end the paper with an analysis of Newcomb's problem, which is far more controversial.

Ultimately, we hope that this work will contribute to resolving ambiguities in decision theory, by offering precise definitions, a common modeling language, and concrete evaluative criteria, thereby facilitating progress in the field.

\section{Notation}

We adhere to standard statistical notation. In particular, we use upper-case letters (e.g., $A$, $U$) for random variables, lower case (e.g., $a$, $u$) for fixed constants (e.g., realizations of random variables), $p(\cdot)$ for probabilities, $E(\cdot)$ for expected values and $|$ for statistical conditioning, e.g., $E(Y|X=x)$ denotes the conditional expected value of $Y$, given that $X$ is equal to $x$. For simplicity, we use $p$ for both probabilities (for discrete variables) and probability densities (for continuous variables). 

\section{The smoking lesion problem}

The smoking lesion problem describes a situation where a population of subjects --  or `agents' -- can choose between two acts: smoking or not smoking. Smoking may be perceived as being pleasant, thus potentially increasing an agent's `utility', However, smoking may also cause cancer, which is certainly unpleasant, thus decreasing the utility.\footnote{In many formulations of the problem, smoking has no causal effect on cancer at all. To keep the example more general, we allow for the possibility that smoking causes cancer, at least for some agents.} The twist of the problem is that some subjects have a genetic lesion that makes them more prone to smoking, while at the same time increasing the risk of cancer, thus inducing a statistical association between smoking and cancer in the population, even if smoking does not cause cancer.   

\begin{figure}[h]
\begin{center}
\[
    \xymatrix{  G \ar[r] \ar@/^2pc/[rr] & A\ar[r] \ar@/_2pc/[rr] & Y \ar[r] & U
  } 
\]
\caption{Causal diagram for the smoking lesion problem.}
	\label{fig:smokinglesion}
\end{center}
\end{figure}
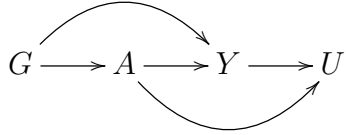

This is a standard example of what statisticians would refer to as `confounding'. Let the binary variables $A$, $Y$ and $G$ denote `smoking', `cancer' and `lesion' respectively (1 for `yes', 0 for `no'). Finally, let $U$ denote `utility'. Then, the situation can be depicted by the causal diagram in Figure \ref{fig:smokinglesion}. The arrows from $G$ to $A$ and $Y$ represent the causal effects of the lesion on smoking and cancer, the arrow from $A$ to $Y$ represents a causal effect of smoking on cancer, and the arrows from $A$ and $Y$ to $U$ represent the causal effects of smoking and cancer on utility. The absence of arrow from $G$ to $U$ reflects an assumption that the lesion has no direct (i.e., not mediated through $A$ or $Y$) causal effect on the utility.  

Although the causal diagram in Figure \ref{fig:smokinglesion} is intuitively appealing, we emphasize that, without further specification, it remains merely a heuristic device -- useful for conveying intuition but insufficient for deriving formal results. In the next section we show how the NPSEM framework provides a rigorous and precise, yet simple, formalization of the diagram.

\section{NPSEMs}

In essence, a NPSEM consists of a vector of `exogenous' random error terms $\boldsymbol{\varepsilon}$ with a distribution $p(\boldsymbol{\varepsilon})$, and a vector of `endogenous' variables that are causally determined by the error terms through a set of functions $\boldsymbol{f}$.\footnote{The terms `endogenous' and `exogenous' essentially mean `explained' and `not explained' by the model. The name `error term' is somewhat of a misnomer, since it suggests that there is something erroneous in the model formulation. However, this name is widely accepted in the literature, so we adopt it here as well.} 

A NPSEM for the causal diagram in Figure \ref{fig:smokinglesion} looks like this:
\begin{eqnarray}
\label{eq:npsemsmoking}
\left . \begin{array}{rcl}
G&:=&f_G(\varepsilon_G)\\
A&:=&f_A(G,\varepsilon_A)\\
Y&:=&f_Y(G,A,\varepsilon_Y)\\
U&:=&f_U(A,Y,\varepsilon_U)
\end{array} \right\},
\end{eqnarray}
where $\boldsymbol{\varepsilon}=(\varepsilon_G,\varepsilon_A,\varepsilon_Y,\varepsilon_U)$ and $\boldsymbol{f}=(f_G,f_A,f_Y,f_U)$. In this model, the error term $\varepsilon_G$ is an abstract construction that represents all factors (e.g., parental genetics) that determine the lesion status. Similarly, the error term $\varepsilon_A$ represents all factors, apart from lesion status, that determine the smoking status, the error term $\varepsilon_Y$ represents all factors, apart from lesion and smoking status, that determine the cancer status, and the error term $\varepsilon_U$ represents all factors, apart from smoking and cancer, that determine the utility.\footnote{In the standard formulation of the smoking lesion problem, the utility is a deterministic function of smoking status and cancer status, e.g., $A=1$ and $Y=0$ always gives a utility of +10. By having an error term $\varepsilon_U$ in the model for $U$, we allow for the more general (and realistic) case where the utility may depend on other factors as well, such as a genetic inclination to find smoking pleasurable.} By convention, the error terms are implicit in the causal diagram corresponding to the NPSEM.

The function $f_G$, specifies how the lesion status is determined by the value of $\varepsilon_G$. Similarly, the function $f_A$ specifies how the smoking status is determined by the lesion status and the value of $\varepsilon_A$, the function $f_Y$ specifies how the cancer status is determined by the lesion status, the smoking status and the value of $\varepsilon_Y$, and the function $f_U$ specifies how the utility is determined by the smoking status, the cancer status and the value of $\varepsilon_U$. These functions can be arbitrarily complex, and are left entirely unspecified; thus, the model is referred to as `nonparametric'. In the causal diagram corresponding to the NPSEM, an arrow from a variable $V_1$ to another variable $V_2$ indicates that $V_1$ appears as an argument in the function $f_{V_2}$ for $V_2$. 

NPSEMs are `structural' (as a synonym for `causal'), in the sense that each equation in the model represents a causal, not statistical, relationship. Thus, in contrast to ordinary equations that are symmetric, structural equations are asymmetric; causality goes from the right-hand side to the left-hand side, not the other way around. We have indicated this by using the assignment operator `$:=$' in the equation system instead of ordinary equality signs. 

The error terms $\boldsymbol{\varepsilon}$ are taken to be statistically independent, so that the joint distribution $p(\boldsymbol{\varepsilon})$ factorizes into a product of marginal distributions. For instance, under the NPSEM in (\ref{eq:npsemsmoking}) for the smoking lesion problem, we have that $p(\varepsilon_G,\varepsilon_A,\varepsilon_Y,\varepsilon_U)=p(\varepsilon_G)p(\varepsilon_A)p(\varepsilon_Y)p(\varepsilon_U)$. This may seem like a very restrictive condition, which would make the model inappropriate for many real situations. For instance, suppose we believe that there are other factors, apart from the lesion, that affect the smoking status $A$ and cancer status $Y$. These factors would presumably be part of $\varepsilon_A$ and $\varepsilon_Y$, thus making them statistically dependent. However, we can handle this by explicitly including all common causes of all pairs of endogenous variables in the model. For instance, we can represent all the common causes of $A$ and $Y$, apart from $G$, by a new endogenous variable that we arbitrarily call $W$. The NPSEM thus becomes
\begin{eqnarray*}
W&:=&f_W(\varepsilon_W)\\
G&:=&f_G(\varepsilon_G)\\
A&:=&f_A(W,G,\varepsilon_A)\\
Y&:=&f_Y(W,G,A,\varepsilon_Y)\\
U&:=&f_U(A,Y,\varepsilon_U)
\end{eqnarray*}
By explicit inclusion of $W$ in the model, it is arguably reasonable to consider the error terms $\varepsilon_A$ and $\varepsilon_Y$ as independent.

\section{Agents}

The term `agent' is central in decision theory. Different agents may act differently when faced with a particular scenario, and the same act may have different consequences (e.g., result in different utilities) for different agents. In a NPSEM, the endogenous variables are entirely determined by the error terms $\boldsymbol{\varepsilon}$, and variation in the endogenous variables is due to the random variation in $\boldsymbol{\varepsilon}$. Thus, in a NPSEM, it is natural to think of an `agent' as a particular instantiation of the error terms $\boldsymbol{\varepsilon}$, and the distribution of the error terms $p(\boldsymbol{\varepsilon})$ as representing the distribution of agents in the population under consideration. 

For instance, under the NPSEM in  (\ref{eq:npsemsmoking}) for the smoking lesion problem, an `agent' is represented by a joint value of $\boldsymbol{\varepsilon}=(\varepsilon_G,\varepsilon_A,\varepsilon_Y,\varepsilon_U)$. For a particular agent $\boldsymbol{\varepsilon}$, the lesion status is 
\begin{eqnarray*}
G(\boldsymbol{\varepsilon}):=f_G(\varepsilon_G)
\end{eqnarray*}
the smoking status is 
\begin{eqnarray*}
A(\boldsymbol{\varepsilon})&:=&f_A(G(\boldsymbol{\varepsilon}),\varepsilon_A)\\
&=&f_A(f_G(\varepsilon_G),\varepsilon_A),
\end{eqnarray*}
the cancer status is  
\begin{eqnarray*}
Y(\boldsymbol{\varepsilon})&:=&f_Y(G(\boldsymbol{\varepsilon}),A(\boldsymbol{\varepsilon}),\varepsilon_Y)\\
&=&f_Y(f_G(\varepsilon_G),f_A(f_G(\varepsilon_G),\varepsilon_A),\varepsilon_Y)
\end{eqnarray*}
and the utility is
\begin{eqnarray*}
U(\boldsymbol{\varepsilon})&:=&f_U(A(\boldsymbol{\varepsilon}),Y(\boldsymbol{\varepsilon}),\varepsilon_U)\\
&=&f_U(f_A(f_G(\varepsilon_G),\varepsilon_A),f_Y(f_G(\varepsilon_G),f_A(f_G(\varepsilon_G),\varepsilon_A),\varepsilon_Y),\varepsilon_U).
\end{eqnarray*}
Here we have used the notation $G(\boldsymbol{\varepsilon})$, $A(\boldsymbol{\varepsilon})$, $Y(\boldsymbol{\varepsilon})$ and $U(\boldsymbol{\varepsilon})$ to explicitly indicate that these variables are functions of $\boldsymbol{\varepsilon}$. When an agent is drawn at random from the population under consideration (i.e., when $\boldsymbol{\varepsilon}$ is drawn at random from $p(\boldsymbol{\varepsilon})$), randomness is propagated to the endogenous variables through the functions $\boldsymbol{f}$. In this case, we often keep the argument $\boldsymbol{\varepsilon}$ implicit. For instance, $U$ refers to the for the utility for an agent drawn at random from the population.

\section{Counterfactuals}

Decision theory often requires us to consider a counterfactual world where an endogenous variable is set by external intervention to a fixed constant. \textcite{lewis1973counterfactuals} referred to the resulting counterfactual scenario as `the closest possible world'. In a NPSEM, this counterfactual scenario is modeled by deleting the equation for the variable that is intervened upon, and setting that variable equal to the fixed constant in all other equations.

For instance, in the smoking lesion problem, suppose that an external intervention sets the smoking status for a particular agent $\boldsymbol{\varepsilon}$ to level $a$. The agent's lesion status is not affected by this intervention, since $G$ is not causally downstream from $A$ in the model. However, their counterfactual cancer status becomes
\begin{eqnarray*}
Y(a,\boldsymbol{\varepsilon}):=f_Y(G(\boldsymbol{\varepsilon}),a,\varepsilon_Y)
\end{eqnarray*}
and their counterfactual utility becomes
\begin{eqnarray*}
U(a,\boldsymbol{\varepsilon})&:=&f_U(a,Y(a,\boldsymbol{\varepsilon}),\varepsilon_U)\\
&=&f_U(a,f_Y(G(\boldsymbol{\varepsilon}),a,\varepsilon_Y),\varepsilon_U).
\end{eqnarray*}

Under this counterfactual model, an intervention that sets an endogenous variable to the value that it would have had, even in the absence of intervention, leaves all other variables unchanged as well. This relation between factual and counterfactual variables is referred to as `consistency' in the modern causal inference literature \parencite{pearl2010consistency}.\footnote{\textcite{gibbard1978counterfactuals} articulated consistency as (page 6): `...if I actually do $a$, then the $a$-world which... is most like the actual world will be the actual world itself'.} To be explicit, for any pair of variables $V_1(\boldsymbol{\varepsilon})$ and $V_2(\boldsymbol{\varepsilon})$, let $V_2(v_1,\boldsymbol{\varepsilon})$ be the counterfactual value of $V_2(\boldsymbol{\varepsilon})$ if $V_1(\boldsymbol{\varepsilon})$ were set to $v_1$. Then consistency states that  
\begin{eqnarray}
\label{ass:consistency}
V_1(\boldsymbol{\varepsilon})=v_1\Rightarrow V_2(v_1,\boldsymbol{\varepsilon})=V_2(\boldsymbol{\varepsilon}).
\end{eqnarray}

As with factual variables, we often keep the argument $\boldsymbol{\varepsilon}$ in counterfactual variables implicit when drawing agents at random from the population. For instance, $U(a)$ refers to the counterfactual utility of an agent drawn at random from the population, if $A$ were set to $a$ by external intervention. 

\section{Statistical associations and causal effects}

To measure the statistical association between an act $A$ and a utility outcome $U$ in a population of agents, we may use the expected value of $U$, conditional on (given) $A$, denoted with $E(U|A)$. For instance, with the binary act `smoking', the mean difference $E(U|A=1)-E(U|A=0)$ measures how the utility differs, on average, between agents who factually smoke and factually don't smoke.  

In contrast, to measure the causal effect of $A$ on $U$, we use the expected value of the counterfactual utility $U(a)$, denoted with $E(U(a))$.\footnote{Another common notation is $E(U|do(a))$, where the `do-operator' $do(\cdot)$ used by \textcite{pearl2009causality} is similar in spirit to the `counterfactual conditional operator' $\square\!\!\!\rightarrow$ proposed by \textcite{lewis1973counterfactuals}.} For instance, with the binary act `smoking', the mean difference $E(U(a=1))-E(U(a=0))$ measures how the utility differs, on average, between the two counterfactual scenarios where all agents smoke and all agents don't smoke.  

In a NPSEM, variation in the endogenous variables is due to the random variation in the exogenous error terms $\boldsymbol{\varepsilon}$, which is propagated through the functions $\boldsymbol{f}$. Thus, expectations such as those in $E(U|A)$ and $E(U(a))$ are taken over the distribution $p(\boldsymbol{\varepsilon})$ of the error terms. In explicit notation, we have that  
\begin{eqnarray}
\label{eq:a1}
E(U|A=a)&=&\sum_ uu\times p(U(\boldsymbol{\varepsilon})=u|A(\boldsymbol{\varepsilon})=a)\nonumber\\
&=&\sum_uu\times \frac{p(U(\boldsymbol{\varepsilon})=u,A(\boldsymbol{\varepsilon})=a)}{p(A(\boldsymbol{\varepsilon})=a)}\nonumber\\
&=&\sum_uu\times\frac{\displaystyle\sum_{\boldsymbol{\varepsilon}:\phantom{1}U(\boldsymbol{\varepsilon})=u, A(\boldsymbol{\varepsilon})=a}p(\boldsymbol{\varepsilon})}{\displaystyle\sum_{\boldsymbol{\varepsilon}:\phantom{1} A(\boldsymbol{\varepsilon})=a}p(\boldsymbol{\varepsilon})}
\end{eqnarray}
and 
\begin{eqnarray}
\label{eq:a2}
E(U(a))&=&\sum_uu\times p(U(a,\boldsymbol{\varepsilon})=u)\nonumber\\
&=&\sum_uu\times\displaystyle\sum_{\varepsilon:\phantom{1}U(a,\boldsymbol{\varepsilon})=u}p(\boldsymbol{\varepsilon}).
\end{eqnarray}

\section{Evidential and causal decision theory}
\label{sec:dec}

As before, let $A$ be an act variable and $U$ a utility variable. With slight variation in notation, modern decision theory \parencite[e.g.,][]{joyce2007,hitchcock2016conditioning,levinstein2020cheating}, defines EDT agents as those agents who act according to the optimization criterion  
\begin{eqnarray}
\label{eq:EDT}
A:=\argmax_a E(U|A=a),
\end{eqnarray}
and CDT agents as those agents who act according to the optimization criterion 
\begin{eqnarray}
\label{eq:CDT}
A:=\argmax_a E(U(a)).
\end{eqnarray}
An issue with these definitions is that many authors emphasize that the expectations in the optimization criteria (\ref{eq:EDT}) and (\ref{eq:CDT}) should be interpreted as subjective models rather than objective truths.\footnote{When we say `objective truth', we mean the expectations in (\ref{eq:a1}) and (\ref{eq:a2}) under the distribution $p(\boldsymbol{\varepsilon})$ for the population of agents under consideration.} However, the notation in (\ref{eq:EDT}) and (\ref{eq:CDT}) does not capture this distinction between subjective and objective expectations.  

To address this issue, we propose to augment the NPSEM with a variable $M$ that contains the agent's subjective model for the objective reality (i.e., for the NPSEM to which the agent belong). We will index quantities in this model with $M$; for instance, $E(U|A=a;M)$ and $E(U(a);M)$ denote the agent's subjective models for $E(U|A=a)$ and $E(U(a))$, respectively.\footnote{This is analogous to standard statistical notation for parametric regression models, e.g., $E(Y|X;\beta)$ is a parametric model for $E(Y|X)$. However, whereas $\beta$ in a regression model is typically assumed to have one fixed value, $M$ has potentially one distinct value for each agent.} It will be useful to also augment the NPSEM with an endogenous variable $D$, which explicitly represents the decision theory that the agent follows.\footnote{Our variable $D$ is similar in spirit to the `decision algorithm variable' $\underline{\text{FDT}(P,G)}$ proposed by \textcite{levinstein2020cheating} and the `decision rule variable' $\widetilde{D}$ proposed by   \textcite{macdermott2023characterisingdecisiontheoriesmechanised}. However, these authors did not use NPSEMs to formalize their notion of a `decision algorithm/rule', and they did not formally distinguish between subjective and objective expectations.}

With these augmentations, we define different decision theories in terms of how an agent's subjective model determines the value of the act variable $A$. \begin{definition}
\label{def:EDTnew}
\emph{Evidential decision theory (EDT)}. 
\begin{eqnarray*}
(D=\text{EDT},M=m)\Rightarrow A:=\argmax_a E(U|A=a;m).
\end{eqnarray*}
\end{definition}
\begin{definition}
\label{def:CDTnew}
\emph{Causal decision theory (CDT).} 
\begin{eqnarray*}
(D=\text{CDT},M=m)\Rightarrow A:=\argmax_a E(U(a);m).
\end{eqnarray*}
\end{definition}

\begin{figure}[h]
\begin{center}
\[
    \xymatrix{  G \ar[rrr] \ar@/^2pc/[rrrr] & && A\ar[r] \ar@/_2pc/[rr] & Y \ar[r] & U\\
    & D \ar[r] \ar[urr]& M \ar[ur]&&&
  } 
\]
\caption{Causal diagram for the smoking lesion problem, augmented with the agent's decision theory $D$ and subjective model $M$.}
	\label{fig:dma}
\end{center}
\end{figure}
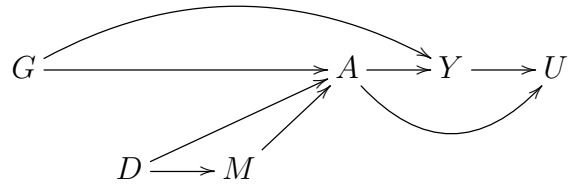

As an example, an augmented NPSEM for the smoking lesion problem is given by
\begin{eqnarray*}
\label{eq:npsemsmokingdm}
G&:=&f_G(\varepsilon_G)\nonumber\\
D&:=&f_D(\varepsilon_D)\nonumber\\
M&:=&f_M(D,\varepsilon_M)\\
A&:=&f_A(D,M,G,\varepsilon_A)\nonumber\\
Y&:=&f_Y(G,A,\varepsilon_Y)\nonumber\\
U&:=&f_U(A,Y,\varepsilon_U)
\end{eqnarray*}
with corresponding causal diagram in Figure \ref{fig:dma}. The absence of arrows from $G$ to $D$ and $M$ in Figure \ref{fig:dma} reflects an assumption that the lesion does not affect the agent's decision theory or the agent's subjective model. This assumption may be reasonable, but it is not required by the model framework \emph{per se}, and it can be relaxed by simply adding $G$ as an argument to the function $f_D$ and/or the function $f_M$. Note that, in order for $G$ to have a causal effect on $A$ (i.e., in order for $f_A$ to be a non-trivial function of $G$), there has to be agents in the population who are neither EDT nor CDT agents. This is because, for EDT and CDT agents, $A$ is completely determined by $D$ and $M$, irrespective of the value of $G$. Thus if all agents were EDT or CDT agents, then $G$ would have no causal effect on $A$. The arrow from $D$ to $M$ reflects that the agent's decision theory may affect the agent's subjective models. For instance, a CDT agent may more carefully contemplate their model for $E(U(a))$ than an EDT agent. 

\section{Personal decision theory}

The expectations in Definitions \ref{def:EDTnew} and \ref{def:CDTnew} are (models for) population averages. For a given agent, one could argue that it makes more sense to attempt to maximize their own utility, rather than maximizing an average utility in the population to which they belong. To accommodate this, we conceive of each agent $\boldsymbol{\varepsilon}$ to have a subjective probabilistic model $p(U(a,\boldsymbol{\varepsilon})|\boldsymbol{\varepsilon};M)$ for their own counterfactual utility $U(a,\boldsymbol{\varepsilon})$ when $A$ is set to $a$, and we let  $E(U(a,\boldsymbol{\varepsilon})|\boldsymbol{\varepsilon};M)$ denote the expected utility under this subjective model. By conditioning on $\boldsymbol{\varepsilon}$ in the expressions  $p(U(a,\boldsymbol{\varepsilon})|\boldsymbol{\varepsilon};M)$ and $E(U(a,\boldsymbol{\varepsilon})|\boldsymbol{\varepsilon};M)$, these apply to the particular agent $\boldsymbol{\varepsilon}$. Note that $U(a,\boldsymbol{\varepsilon})$ is a deterministic function of $a$ and $\boldsymbol{\varepsilon}$. Thus, if the agent's subjective model $p(U(a,\boldsymbol{\varepsilon})|\boldsymbol{\varepsilon};M)$ is correct, then it is a degenerate distribution with probability mass 1 at, and expected value equal to, the true value of $U(a,\boldsymbol{\varepsilon})$.

In this notation, we define \emph{personal decision theory} (PDT) as maximizing the agent's own expected counterfactual utility:
\begin{definition}
\label{def:PDTnew}
\emph{Personal decision theory (PDT)}. 
\begin{eqnarray*}
(D=\text{PDT},M=m)\Rightarrow A:=\argmax_a E(U(a,\boldsymbol{\varepsilon})|\boldsymbol{\varepsilon};m).
\end{eqnarray*}
\end{definition}

Intuitively, one would expect PDT to outperform CDT (and EDT) if there is strong heterogeneity in the causal effect of $A$ on $U$, so that a particular act gives high positive utility for some agents, but high negative utility for other agents. We will see an example of this in the next section. 

Given the intuitive appeal of PDT -- after all, why would an agent care about anything other than their own utility? -- we suspect that some authors may have implicitly interpreted CDT, and perhaps also EDT, as maximizing a subjective model of an individual agent’s utility, rather than a population-level expectation. However, we have found no clear statements to this effect in the existing literature, nor any formal distinction between these two interpretations. Furthermore, at the agent-level it makes little sense to distinguish causal from evidential utility. This is because, under consistency (\ref{ass:consistency}), the agent's utility under the act $A=a$ is the same, regardless of whether the act results from an external intervention or from the agent's own choice.\footnote{In principle, one could of course conceive of the agent having one subjective expectation $E(U(a,\boldsymbol{\varepsilon})|\boldsymbol{\varepsilon};M)$ and another subjective expectation  $E(U(\boldsymbol{\varepsilon})|A(\boldsymbol{\varepsilon})=a,\boldsymbol{\varepsilon};M)$. However, if these are correct, then they are identical to each other and to the true value of $U(a,\boldsymbol{\varepsilon})$.} 

\section{An objective performance metric}

Determining which decision theory is `best' requires an objective performance metric. One might be tempted to argue that EDT implicitly uses the metric $E(U|A=a)$, while CDT uses 
$E(U(a))$. However, this argument is unsatisfactory for two reasons. First, these metrics evaluate the expected utility under different acts, but they do not reference the decision theories that generate those acts. As such, they allow for comparison between acts, but not between the underlying theories themselves. Second, a meaningful comparison between decision theories requires that they be evaluated using a common performance metric -- one that applies equally across all theories under consideration, regardless of their internal reasoning processes. 

The appropriate choice of performance metric depends on the perspective we adopt. We will consider the viewpoint of a policy-maker -- such as a legislator or public official -- who is contemplating the possibility of enforcing a particular decision theory across the population through intervention. In other words, the policy-maker asks: \emph{If I could influence how people make decisions, for example through education or regulation, which decision theory should I promote in order to maximize overall expected utility?} This question naturally leads to the performance metric $E(U(d))$, where $U(d)$ denotes the counterfactual utility that would result if the population’s decision theory $D$ were set to $d$.

\section{Conditions for optimality of PDT}

Without additional assumptions, it is not possible to determine which of EDT, CDT and PDT is best according to the performance metric $E(U(d))$. There are two reasons for this. First, these theories are defined in terms of agents' subjective models, which may not correspond (equally) well to the objective reality. Second, the decision theory that an agent follows may have direct causal effects on their utility, not mediated through the act implied by the theory. For instance, it could be that adhering to CDT makes the agent excessively introspective, thus decreasing the agent's mental well-being, even though the act implied by CDT may have a stronger positive causal effect on the agent's mental well-being than the act implied by EDT.  

However, if the agents' subjective models are correct, and there are no direct causal effects of the decision theory on the utility, then, with respect to the performance metric $E(U(d))$, it can be shown that CDT is better than EDT, and that PDT is optimal among all decision theories. We state these assumptions and results formally below. Theorems \ref{theo:const}, \ref{theo:optCDT} and \ref{theo:optPDT} are proven in the Appendices \ref{app:const}, \ref{app:optCDT} and \ref{app:optPDT}, respectively.

\begin{assumption}
\label{ass:accuracy}
\emph{Correct subjective models.} Let $M(d)$ be the counterfactual subjective model if $D$ were set to $d$, and $U(a)$ be the counterfactual utility if $A$ were set to $a$. Then, for all $(a,m)$, we assume 
\begin{eqnarray*}
E(U|A=a;M(\text{EDT})=m)&=&E(U|A=a)\\
E(U(a);M(\text{CDT})=m)&=&E(U(a))\\
E(U(a,\boldsymbol{\varepsilon})|\boldsymbol{\varepsilon};M(\text{PDT})=m)&=&U(a,\boldsymbol{\varepsilon})\text{ for all }\boldsymbol{\varepsilon}
\end{eqnarray*}
\end{assumption}
Note that Assumption \ref{ass:accuracy} does not state that \emph{all} subjective models are \emph{always} correct, only the models that are required to determine the act under each respective decision theory. For instance, the assumption states that the subjective model $E(U|A=a;M(\text{EDT}))$ is identical to the true expectation $E(U|A=a)$, but not that this also holds for the subjective models $E(U|A=a;M(\text{\text{CDT}}))$ and $E(U|A=a;M(\text{\text{PDT}}))$. This is important since the model required to determine the act under a particular decision theory may be more carefully contemplated by the agent (i.e., more likely to be correct) than other models, e.g., $E(U|A=a;M(\text{\text{EDT}}))$ is more likely to be correct than $E(U|A=a;M(\text{\text{CDT}}))$ and $E(U|A=a;M(\text{\text{PDT}}))$. 
\begin{assumption}
\label{ass:nodirect}
\emph{No direct causal effect of decision theory on utility}. Let $U(a)$ be the counterfactual utility if $A$ were set to $a$, and let $U(a,d)$ be the counterfactual utility if $A$ and $D$ were set to $a$ and $d$, respectively. Then we assume
\begin{eqnarray*}
U(a,d)=U(a)\text{ for all }a,d.
\end{eqnarray*}
\end{assumption}
Assumption \ref{ass:nodirect} holds under the causal diagram in Figure \ref{fig:dma} for the smoking lesion problem, since all causal paths from $D$ to $U$ are intersected by $A$. Note that the assumption would still hold if we would add arrows from $G$ to $D$ and $M$. We will discuss possible violations of this assumption in Section \ref{sec:newcomb}, for Newcomb's problem.

\begin{theorem}
\label{theo:const}
Let $A(d)$ be the counterfactual act if $D$ were set to $d$. Under Assumption \ref{ass:accuracy}, $A(\text{EDT})$ and $A(\text{CDT})$ are constant across all agents, i.e., $A(\text{EDT},\boldsymbol{\varepsilon})$ and $A(\text{CDT},\boldsymbol{\varepsilon})$ do not depend on $\boldsymbol{\varepsilon}$. 
\end{theorem}
Theorem \ref{theo:const} is intuitively reasonable -- if all agents have correct subjective models for the population to which they belong, then they should also choose the same act under decision theories that only depend on population quantities, like EDT and CDT.

\begin{theorem}
\label{theo:optCDT}
Let $U(d)$ and $A(d)$ be the counterfactual utility and act, respectively, if $D$ were set to $d$. Under Assumptions \ref{ass:accuracy} and \ref{ass:nodirect}, we have that
\begin{eqnarray*}
E(U(\text{CDT}))\geq E(U(d))\text{ for all }d\text{ such that }A(d)\text{ is constant}.
\end{eqnarray*}
\end{theorem}
An immediate corollary of Theorems \ref{theo:const} and \ref{theo:optCDT} is that, under Assumptions \ref{ass:accuracy} and \ref{ass:nodirect}, CDT beats EDT with respect to the performance metric $E(U(d))$.

\begin{theorem}
\label{theo:optPDT}
Let $U(d)$ denote the counterfactual utility if $D$ were set to $d$. Under Assumptions \ref{ass:accuracy} and \ref{ass:nodirect}, we have that
\begin{eqnarray*}
E(U(\text{PDT}))\geq E(U(d))\text{ for all }d.
\end{eqnarray*}
\end{theorem} 
An immediate corollary of Theorem \ref{theo:optPDT} is that, under Assumptions \ref{ass:accuracy} and \ref{ass:nodirect}, PDT beats CDT with respect to the performance metric $E(U(d))$. To see how this can happen, consider again the smoking lesion problem. Let $U(a)$ be a binary (0=low, 1=high) utility if smoking were set to $a$, and $p_{ij}$ be the proportion of agents in the population with $U(0)=i$ and $U(1)=j$:
\begin{eqnarray*}
p_{ij}=p(U(0)=i, U(1)=j)\text{ for }(i,j)\in\{0,1\}.
\end{eqnarray*}
For simplicity, we assume that there are no agents who are indifferent to smoking, that is, $p_{00}=p_{11}=0$. In this notation, the expected utility if no-one smokes is $E(U(0))=p_{01}$ and the expected utility if everyone smokes is $E(U(1))=p_{10}$. Suppose that $p_{10}>p_{01}$ so that $E(U(1))>E(U(0))$. Under Assumption \ref{ass:accuracy}, CDT agents would correctly identify this relationship so that, if all agents were CDT agents, then all agents would decide to smoke, and the expected utility would equal $p_{10}$. However, if instead all agents were PDT agents, then those agents with $U(0)>U(1)$ would decide not to smoke, so that the expected utility would instead equal $p_{10}+p_{01}=1$. That is, PDT beats CDT by tailoring the act to maximize the utility for each agent, whereas CDT requires each agent to take the act that is best `on average' in the population, but not necessarily for themselves.   

We end this section with a technical remark. The statement $E(U|A=a;M(\text{EDT}))=E(U|A=a)$ in Assumption \ref{ass:accuracy} has a `touch of circularity' to it. To see this, note that, when combined with Definition \ref{def:EDTnew}, it reads 
\begin{eqnarray*}
A&:=&\argmax_a E(U|A=a;M(\text{EDT}))\\
&=&\argmax_a E(U|A=a).
\end{eqnarray*}
Thus, it may appear as if the act $A$ is determined by the expectation $E(U|A=a)$, but the conditioning on $A=a$ in this expectation presupposes that $A$ is already determined.\footnote{The corresponding statements $E(U(a);M(\text{CDT}))=E(U(a))$ and $U(a;M(\text{PDT}))=U(a)$ do not have this issue, since the counterfactual $U(a)$ is well-defined (e.g., `exists') before $A$ is factually determined.} The resolution to this apparent paradox lies in the direction of causality between the model $E(U|A=a;M)$ and the reality $E(U|A=a)$. The model exists (e.g., in the `mind of the agent') before $A$ is determined, and is used to causally determine the value of $A$. Then, when $A$ is eventually determined for all agents, the factual reality may turn out to coincide with the model that generated it. Thus, it is the subjective model that affects the reality, not the other way around.  

\section{Newcomb's problem}
\label{sec:newcomb}

Among all classic problems in the decision theory literature, Newcomb's problem appears to be one of the most controversial \parencite[e.g.,][]{gibbard1978counterfactuals,joyce2007,hitchcock2016conditioning,bales2018richness,levinstein2020cheating}. Notably, many decision theorists have expressed strong opinions about which decision theory is `best' for this problem, without specifying an objective performance metric with which to rank different theories. In this section, we will propose a NPSEMs for Newcomb's problem, and compare EDT, CDT and PDT with respect to the performance metric $E(U(d))$ under the proposed model. 

\subsection{English problem formulation} 

Newcomb's problem (also called Newcomb's paradox) was first published by \textcite{nozick1969newcomb}, and goes like this:\footnote{The exact wording of the problem varies in the literature; in particular, Nozick's original formulation was quite verbose. We believe that our more terse formulation captures the essence of the problem, as perceived by most authors.}\\
\\
\emph{There are two boxes. The agent will get box 1 no matter what, but can decide whether or not to take box 2, which contains a small sum of money. A highly accurate predictor has predicted the agent's choice, and has put a large sum of money in box 1 if and only if they have predicted that the agent will not take box 2. The agent is aware of the setup, but does not know which prediction has been made. Should the agent one-box or two-box?}\\
\\
\textcite{nozick1969newcomb} argued that there are two, seemingly valid, ways to reason about the problem. On the one hand, if the agent takes both boxes, then this will most likely have been predicted, so that box 1 is empty and the agent only gains the small sum of money contained in box 2. However, if the agent takes box 1 only, then this will most likely also have been predicted, so that it contains the large sum of money. Hence, according to this argument, the agent should take box 1 only. On the other hand, the prediction of the agent's choice has already been made and the money has already been put accordingly in the boxes. If the agent takes box 1 only, then the agent only gains what is already put in box 1, whereas if the agent takes both boxes, then the agent gains what is already put in box 1 plus what is already put in box 2. Hence, according to this argument, the agent should take both boxes. These two conclusions are contradicting, which suggests a paradox.

\subsection{A NPSEM for the problem}

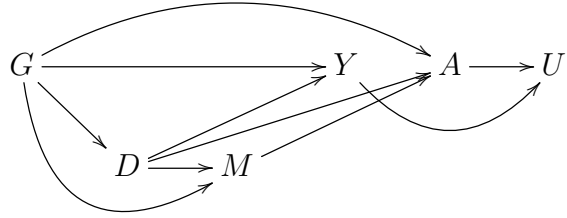
\begin{figure}[h]
\begin{center}
\[
    \xymatrix{  G \ar[rrr] \ar@/^2pc/[rrrr] \ar[dr] \ar@/_3pc/[drr] & && Y \ar@/_2pc/[rr] & A \ar[r] & U\\
    & D \ar[r] \ar[urr] \ar[urrr]& M \ar[urr]&&&
  } 
\]
\caption{Causal diagram for Newcomb's problem.}
	\label{fig:newcomb}
\end{center}
\end{figure}

Let $A$ be the act taken by the agent ($A=0$ for one-boxing, $A=1$ for two-boxing), $Y$ be the prediction of the agent's act ($Y=0$ for predicted to one-box, $Y=1$ for predicted to two-box), $D$ be the agent's decision theory, $M$ be the agent's subjective model, and $U$ be the agent's utility (i.e., the amount of money acquired by the agent), and $G$ be the set of all factors, apart from the agent's decision theory, that the predictor uses to make their prediction (e.g., age, sex, education). Finally, let $x$ be the small sum of money in box 2, and $z$ be the large sum of money put in box 1 if the agent is predicted not to take box 2. According to our reading, a reasonable NPSEM for the problem is given by 
\begin{eqnarray}
\label{eq:newcomb}
\left . \begin{array}{rcl}
G&:=&f_G(\varepsilon_G)\\
D&:=&f_D(G,\varepsilon_D)\\
M&:=&f_M(G,D,\varepsilon_M)\\
Y&:=&f_Y(G,D,\varepsilon_Y)\\
A&:=&f_A(D,M,G,\varepsilon_A)\\
U&:=&xA+z(1-Y)
\end{array}\right\}
\end{eqnarray}
This NPSEM corresponds to the causal diagram in Figure \ref{fig:newcomb}. 

This model has some similarities with that for the smoking lesion problem in Figure \ref{fig:dma}, but also some important structural differences. First, by including arrows from $G$ to $D$ and $M$, we allow the set of predictors to affect the agent's decision theory and subjective model, which we believe is consistent with the standard formulation of Newcomb's problem. Second, by not including an arrow from $A$ to $Y$, we rule out a causal effect of the agent's act on the prediction of the act. Such a causal effect is logically impossible since the prediction precedes the act in time. We have also not included an arrow from $Y$ to $A$, since the agent is unaware of the prediction made and can therefore not be influenced by it. 

Third -- and most importantly -- we have included an arrow from $D$ to $Y$ to capture the possibility that the agent’s decision theory influences the prediction, which we believe is consistent with the standard formulation of the problem. Such effect would be present if, for instance, the predictor is not only able to measure factors that could affect the agent's decision theory (e.g., age, sex and education), but also factors that could be affected by the agent's decision theory (e.g., the agent's eye movements, body language and neural activity). Note that, if $D$ has a causal effect on $Y$, then the predictor can adapt to an external intervention that manipulates the agent's decision theory, so that their prediction may remain accurate under this counterfactual scenario. In contrast, if $D$ has no causal effect on $Y$, then the predictor will be not be able to adapt in this way, and their prediction may no longer remain accurate under such intervention. Thus, not surprisingly, we will see that the presence and magnitude of the causal effect of $D$ on $Y$ is crucial for the comparison of different decision theories. 

When there is a causal effect of $D$ on $Y$, there is also a direct (i.e., not mediated through $A$) causal effect of $D$ on $U$, through the path $D\rightarrow Y\rightarrow U$. This violates Assumption \ref{ass:nodirect}, and as a result, Theorem \ref{theo:optPDT} does not apply; there is no guarantee that PDT is optimal under this model. Nonetheless, the model remains consistent with Assumption \ref{ass:accuracy}, which states that agents' subjective models are accurate. For simplicity, we therefore restrict our analysis to cases where Assumption \ref{ass:accuracy} holds.

\subsection{Performance of different decision theories}

Let $A(d)$, $Y(d)$ and $U(d)$ be the counterfactual act, prediction and utility, respectively, if $D$ were set to $d$. We have that
\begin{eqnarray}
\label{eq:urel}
E(U(d))=xp(A(d)=1)+z(1-p(Y(d)=1)).
\end{eqnarray}
Thus, to compare different decision theories with respect to the performance metric $E(U(d))$, we need to determine $p(A(d)=1)$ and $p(Y(d)=1)$ for the decision theories $d$ under consideration.

We start with $p(A(d)=1)$. Under the NPSEM in (\ref{eq:newcomb}) it is easy to show (see Appendix \ref{app:newcomb}) that, if the decision theory were set to $\text{CDT}$ or $\text{EDT}$, then the agent would two-box:
\begin{eqnarray}
\label{eq:AdCDTPDT}
p(A(\text{CDT})=1)=p(A(\text{PDT})=1)=1.
\end{eqnarray}
If, in addition, the predictor is `sufficiently accurate' (in a specific sense that we define in Appendix \ref{app:newcomb}) and the decision theory were set to $\text{EDT}$, then the agent would one-box:
\begin{eqnarray}
\label{eq:AdEDT}
p(A(\text{EDT})=1)=0.
\end{eqnarray}
We proceed by assuming that this is the case.

We next consider $p(Y(d)=1)$. Given that CDT and PDT dictate two-boxing whereas EDT dictates one-boxing, it seems reasonable to assume that the highly accurate predictor is more likely to predict two-boxing if the decision theory were set to CDT or PDT, than if it were set to EDT, i.e., $p(Y(\text{CDT})=1)>p(Y(\text{EDT})=1)$ and $p(Y(\text{PDT})=1)>p(Y(\text{EDT})=1)$. However, it is not immediately obvious how to rank $p(Y(\text{CDT})=1)$ and $p(Y(\text{PDT})=1)$. In principle we could have that, even though CDT and PDT dictate the same act, CDT agents tend to be more `opaque' to the predictor than PDT agents, so that $p(Y(\text{PDT})=1)>p(Y(\text{CDT})=1)$, or the other way around. However, we proceed by assuming that this is not the case. Specifically, we assume that, if two decision theories $d$ and $d'$ dictate the same constant act for all agents, then the probability of predicting two-boxing is also the same under the counterfactual scenarios when the decision theory is set to either $d$ or $d'$:
\begin{eqnarray}
A(d)=A(d')\Rightarrow p(Y(d)=1)=p(Y(d')=1)\text{ for all }d,d'.
\label{eq:assequal}
\end{eqnarray}
Note that this assumption trivially holds in the special case when $D$ has no effect on $Y$ (i.e., when the arrow from $D$ to $Y$ is absent in Figure \ref{fig:newcomb}), since, in the absence of this effect, $p(Y(d)=1)=p(Y(d')=1)=p(Y=1)$ for all $d,d'$. Under this assumption, CDT and PDT perform equally well: $E(U(\text{CDT}))=E(U(\text{PDT}))$. To compare these with EDT, let
\begin{eqnarray*}
\Delta=p(Y(\text{CDT})=1)-p(Y(\text{EDT})=1)
\end{eqnarray*}
denote the causal effect on $Y$ of setting $D$ to $\text{EDT}$ vs $\text{CDT}$. From (\ref{eq:urel}), (\ref{eq:AdCDTPDT}) and (\ref{eq:AdEDT}) it follows that $E(U(\text{CDT}))>E(U(\text{EDT}))$ if 
\begin{eqnarray*}
\Delta<x/z.
\end{eqnarray*}
Thus, whether or not CDT (and PDT) beats EDT in Newcomb's problem depends on the causal effect $\Delta$ and ratio of money $x/z$ put in the boxes: if $\Delta$ is sufficiently small, relative to $x/z$, then CDT beats EDT. At one extreme, the agent's decision theory entirely determines the prediction, i.e., $Y(\text{CDT})=1-Y(\text{EDT})=1$. In this case,  $\Delta=1>x/z$, so that EDT is better than CDT. At the other extreme, the agent's decision theory has no causal effect on the prediction, i.e., $Y(\text{CDT})=Y(\text{EDT})=Y$. This corresponds to the arrow from $D$ to $Y$ in Figure \ref{fig:newcomb} being entirely absent. In this case, $\Delta=0<x/z$, so that CDT is better than EDT. 

So far, we have only compared EDT, CDT and PDT. Under the assumption in (\ref{eq:assequal}), it is easy to see that $E(U(d))=E(U(\text{EDT}))$ for all decision theories $d$ such that $A(d)=0$, and $E(U(d))=E(U(\text{CDT}))$ for all decision theories $d$ such that $A(d)=1$. Thus, the conclusions above cover all decision theories that are deterministic, in the sense that they dictate a constant act (one-boxing or two-boxing) for all agents. However, these could still be outperformed by a stochastic decision theory. We discuss this in detail in Appendix \ref{app:newcomb2}.  

\section{Discussion}

In this paper we have demonstrated how to use NPSEMs to formalize decision theory. Specifically, we have shown how to formally represent agents within the NPSEM framework, and how to provide unambiguous definitions of EDT and CDT. We have further proposed a novel decision theory that we labeled `personal decision theory' (PDT). We have proposed an objective performance metric that captures the notion of enforcing a decision theory uniformly on a population of agents and we have demonstrated that, with respect to this performance metric, PDT is optimal under certain assumptions. Finally, we have analyzed Newcomb's problem within the proposed framework. 

In line with much of previous literature, we have focused on relatively simple and stylized settings, where agents can only make a single binary decision (e.g., smoke or not smoke, one-box or two-box), and where they act independently of each other. Real decision situations often involve multidimensional decisions that are made repeatedly over time, as well as complex agent-agent interactions. Extending the NPSEM framework to such situations is an important topic for future research.     

While we hope that our proposed framework helps clarify long-standing ambiguities in decision theory, we do not expect it to gain universal acceptance -- especially given the deep-rooted disagreements within the field. Nonetheless, we would urge critics to offer alternative frameworks that can match the key features of ours: an unambiguous definition of the term `agent', a principled separation between subjective and objective elements, a clear distinction between CDT and PDT, and an objective performance metric by which different decision theories can be evaluated. Without such components, we believe it is difficult to achieve meaningful progress in the foundations of decision theory.

\printbibliography

\appendix

\section{Proof of Theorem \ref{theo:const}}
\label{app:const}

We have that  
\begin{eqnarray*}
A(\text{EDT})&=&A(\text{EDT};M(\text{EDT}))\\
&=&\argmax_a E(U|A=a;M(\text{EDT}))\\
&=&\argmax_a E(U|A=a)\\
&=&\text{constant},
\end{eqnarray*}
where the first equality follows from consistency (\ref{ass:consistency}), the second from Definition \ref{def:EDTnew}, and the third from Assumption \ref{ass:accuracy}. The proof for $D=\text{CDT}$ is analogous.

\section{Proof of Theorem \ref{theo:optCDT}}
\label{app:optCDT}

For all $d$, we have that 
\begin{eqnarray}
\label{eq:app1}
E(U(d))&=&E(U(d,A(d,M(d))))\nonumber\\
&=&E(U(A(d,M(d)))),
\end{eqnarray}
where the first equality follows from consistency (\ref{ass:consistency}) and the second from Assumption \ref{ass:nodirect}. We now have that 
\begin{eqnarray*}
&&E(U(\text{CDT}))-E(U(d))\\
&&\phantom{1}=E(U(A(\text{CDT},M(\text{CDT}))))-E(U(A(d,M(d))))\\
&&\phantom{1}=\max_a E(U(a))-E(U(A(d,M(d))))\\
&&\phantom{1}\geq 0,
\end{eqnarray*}
where the first equality follows from (\ref{eq:app1}), the second by combining Assumption \ref{ass:accuracy} with Definition \ref{def:CDTnew}, and the inequality holds if $d$ is such that $A(d,M(d))=A(d)$ is constant. 

\section{Proof of Theorem \ref{theo:optPDT}}
\label{app:optPDT}

We have that 
\begin{eqnarray*}
&&E(U(\text{PDT}))-E(U(d))\\
&&\phantom{1}=E(U(A(\text{PDT},M(\text{PDT}))))-E(U(A(d,M(d))))\\
&&\phantom{1}=E(U(A(\text{PDT},M(\text{PDT})))-U(A(d,M(d))))\\
&&\phantom{1}=E(\max_a U(a,\boldsymbol{\varepsilon})-U(A(d,M(d)),\boldsymbol{\varepsilon}))\\
&&\phantom{1}\geq 0,
\end{eqnarray*}
where the first equality follows from (\ref{eq:app1}), the third by combining Assumption \ref{ass:accuracy} with Definition \ref{def:PDTnew}, and the inequality holds for all $d$.

\section{$A(d)$ for Newcomb's problem}
\label{app:newcomb}

Under the NPSEM in (\ref{eq:newcomb}), we have that 
\begin{eqnarray}
\label{eq:ya}
Y(a)=Y\text{ for all }a.
\end{eqnarray}
We thus have that 
\begin{eqnarray}
\label{eq:ua}
U(a)&=&xa+z(1-Y(a))\nonumber\\
&=&xa+z(1-Y),
\end{eqnarray}
where the second equality follows from (\ref{eq:ya}). The expression in (\ref{eq:ua}) is trivially maximized for $a=1$.  Hence, $A(\text{PDT})=1$. Next, from (\ref{eq:ua}) we have that 
\begin{eqnarray*}
E(U(0))=z(1-p(Y=1))
\end{eqnarray*}
and 
\begin{eqnarray*}
E(U(1))=x+z(1-p(Y=1)).
\end{eqnarray*}
Hence, $E(U(1))>E(U(0))$, so $A(\text{CDT})=1$. Finally, we have that 
\begin{eqnarray*}
E(U|A=0)=z(1-p(Y=0|A=0))
\end{eqnarray*}
and
\begin{eqnarray*}
E(U|A=1)=x+z(1-p(Y=1|A=1)),
\end{eqnarray*}
so that $E(U|A=0)>E(U|A=1)$ if 
\begin{eqnarray}
\label{eq:youden}
p(Y=1|A=1)-p(Y=1|A=0)>x/z.
\end{eqnarray}
Provided that the predictor is more accurate than blind guessing (i.e., $p(Y=1|A=1)-p(Y=1|A=0)>0$), the relation in (\ref{eq:youden}) can always be made to hold by letting $z$ be sufficiently large, in relation to $x$, so that $A(\text{EDT})=0$.  

\section{Stochastic decision theories in Newcomb's problem}
\label{app:newcomb2}

Consider a stochastic decision theory $d$ that dictates the agent to two-box with probability $p(d)$, so that $p(A(d)=1)=p(d)$. Suppose, for the sake of argument, that a stronger version of the assumption in (\ref{eq:assequal}) holds, namely that 
\begin{eqnarray}
\label{eq:assequal2}
p(A(d)=1)=p(A(d'))\Rightarrow p(Y(d)=1)=p(Y(d')=1)=q(p(d))\text{ for all }d,d'.\nonumber\\
\end{eqnarray} 
We then have that 
\begin{eqnarray*}
E(U(d))=xp(d)+z(1-q(p(d)))
\end{eqnarray*}
and
\begin{eqnarray*}
\frac{\partial}{\partial p}E(U(d))=x-z\frac{\partial}{\partial p}q(p(d)).
\end{eqnarray*}
If $\frac{\partial}{\partial p}q(p(d))<x/z$ for all $p$, then $\frac{\partial}{\partial p}E(U(d))>0$ for all $p$, so that the best stochastic decision theory is that which sets $p(d)=1$, i.e., CDT. In contrast, if $\frac{\partial}{\partial p}q(p(d))>x/z$ for all $p$, then $\frac{\partial}{\partial p}E(U(d))<0$ for all $p$, so that the best stochastic decision theory is that which sets $p(d)=0$, i.e., EDT. Between these extremes, the best decision theory is the stochastic decision theory that solves the equation
\begin{eqnarray*}
\frac{\partial}{\partial p}q(p(d))=x/z,
\end{eqnarray*}
so that $\frac{\partial}{\partial p}E(U(d))=0$.
\end{document}